\documentclass[10pt,twocolumn,letterpaper]{article}

\usepackage{wacv}
\usepackage{times}
\usepackage{epsfig}
\usepackage{graphicx}
\usepackage{amsmath}
\usepackage{amssymb}

\usepackage{booktabs}
\usepackage{algorithm}
\usepackage[]{algpseudocode}
\usepackage{pifont}
\newcommand{\cmark}{\ding{51}}
\newcommand{\xmark}{\ding{55}}
\usepackage{stackengine}
\usepackage{pifont}
\def\wacvPaperID{1063} % Enter the WACV Paper ID here

\wacvfinalcopy % *** Uncomment this line for the final submission

\ifwacvfinal
\fi

\ifwacvfinal
\usepackage[breaklinks=true,bookmarks=false]{hyperref}
\else
\usepackage[pagebackref=true,breaklinks=true,colorlinks,bookmarks=false]{hyperref}
\fi

\begin{document}

%%%%%%%%% TITLE
\title{AE-StyleGAN: Improved Training of Style-Based Auto-Encoders}

% \and

\author{
Ligong Han\thanks{Equal contribution.}~~,\textsuperscript{\rm 1}
Sri Harsha Musunuri$^*$,\textsuperscript{\rm 1}
Martin Renqiang Min,\textsuperscript{\rm 2}\\
Ruijiang Gao,\textsuperscript{\rm 3}
Yu Tian,\textsuperscript{\rm 1}
Dimitris Metaxas\textsuperscript{\rm 1}\\
\textsuperscript{\rm 1}Rutgers University~
\textsuperscript{\rm 2}NEC Labs America
\textsuperscript{\rm 3}The University of Texas at Austin\\
lh599@rutgers.edu, harsha.musunuri@rutgers.edu, ~renqiang@nec-labs.com\\ ruijiang@utexas.edu, ~yt219@cs.rutgers.edu, ~dnm@cs.rutgers.edu
}

\maketitle
\ifwacvfinal
\thispagestyle{empty}
\fi

%%%%%%%%% ABSTRACT
\begin{abstract}
    %StyleGANs have shown impressive results  on generation and manipulation in recent years, thanks to its disentangled style latent space. A lot of efforts have been made in inverting a {\em pretrained} generator, where an encoder is trained ad hoc after the generator is trained in a two-stage fashion. In this paper, we ask a scientific question: how to train a style-based {\em autoencoder} where the encoder and generator is optimized {\em end-to-end}.
    StyleGANs have shown impressive results on data generation and manipulation in recent years, thanks to its disentangled style latent space. A lot of efforts have been made in inverting a {\em pretrained} generator, where an encoder is trained ad hoc after the generator is trained in a two-stage fashion. In this paper, we focus on style-based generators asking a scientific question: Does forcing such a generator to reconstruct real data lead to more disentangled latent space and make the inversion process from image to latent space easy? We describe a new methodology to train a style-based {\em autoencoder} where the encoder and generator are optimized {\em end-to-end}. We show that our proposed model consistently outperforms baselines in terms of image inversion and generation quality. Supplementary, code, and pretrained models are available on the project website\footnote{\url{https://github.com/phymhan/stylegan2-pytorch}}. %Supplementary is provided in here\footnote{\url{https://www.dropbox.com/s/6a4odr09mxgf6z8/AE-StyleGAN_supp.pdf?dl=0}}.
\end{abstract}

%%%%%%%%% BODY TEXT
\section{Introduction}
\begin{figure*}[t]
\begin{center}
% \fbox{\rule{0pt}{2in} \rule{.9\linewidth}{0pt}}
\includegraphics[width=1\linewidth]{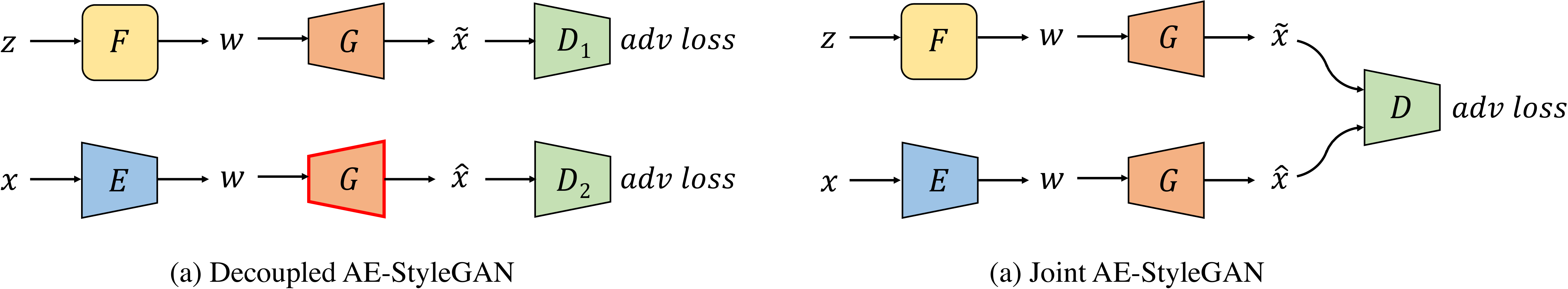}
\end{center}
   \caption{\textbf{Model Architecture.} \textbf{(a) Decoupled AE-StyleGAN}: $E$, $F$, $G$, $D$ are encoder, multi-layer perceptron, generator, and discriminator, respectively. We learn $E$ to map image space to $\mathcal{W}$ (or $\mathcal{W}^+$) space. The red outline around $G$ is to tell that in the encoder update step, generator is in gradient-freeze stage and only the encoder gets updated. \textbf{(b) Joint AE-StyleGAN}: Here, we train encoder along with generator at the same time with a shared discriminator. Please refer Algorithm~\ref{alg:aegan_decouple} and Algorithm~\ref{alg:aegan_joint} for more details.}
\label{fig:arch}
\end{figure*}
% \ligong{the space between left and right can be smaller like 1/2 or 2/3 of the original space.}
The generative adversarial networks (GAN)~\cite{goodfellow2014generative} in deep learning estimates how data points are generated in a probabilistic framework. It consists of two interacting neural networks: a generator, $G$, and a discriminator, $D$, which are trained jointly through an adversarial process. The objective of $G$ is to synthesize fake data that resemble real data, while the objective of $D$ is to distinguish between real and fake data. Through an adversarial training process, the generator can generate fake data that match the real data distribution. GANs have been applied to numerous tasks ranging from conditional image synthesis~\cite{mirza2014conditional,han2020robust,han2020unbiased,Han_2021_ICCV}, image translation~\cite{isola2017image,zhu2017unpaired,han2018learning,park2019semantic,DBLP:journals/corr/abs-1804-04732}, text-to-image generation~\cite{Zhang2016StackGANTT,reed2016generative}, image restoration~\cite{ledig2017photo,nazeri2018image,afifi2021histogan} \etc

Most notably, StyleGANs~\cite{karras2019style,karras2020analyzing} propose a novel style-based generator architecture and attain state-of-the-art visual quality on high-resolution images. As it can effectively encode rich semantic information in the latent space~\cite{DBLP:journals/corr/abs-2004-14367,DBLP:journals/corr/abs-1907-10786}, we can edit the latent code and synthesize images with various attributes, such as aging, expression, and light direction. However, such manipulations in the latent space are only applicable to images generated from GANs rather than to any given real images due to the lack of inference functionality or the encoder in GANs.

In contrast, GAN inversion aims to invert a given image back into the latent space of a {\em pretrained} generator. The image can then be faithfully reconstructed from the inverted code by the generator. GAN inversion enables the controllable directions found in latent spaces of the existing trained GANs to be applicable to real image editing, without requiring {\it ad-hoc} supervision or expensive optimization. As StyleGAN is known to have a disentangled latent space which offers control and editing capabilities and has become common practice~\cite{abdal2019image2stylegan,zhu2020domain,DBLP:journals/corr/abs-1911-11544,DBLP:journals/corr/abs-2008-02401} to encode real images into an extended latent space, $\mathcal{W}^+$, defined by the concatenation of `n' different 512-dimensional $w$ vectors (or {\em styles}). 

In this paper, we focus on a scientific question of how to train a style encoder along with a style-based generator. This essentially requires training a {\em style-based autoencoder} networks where the encoder and generator are optimized simultaneously {\em end-to-end}. This is different from GAN inversion literature where $G$ is fixed and autoencoding is implemented stage-wise. Adversarial Latent Autoencoders (or ALAE~\cite{pidhorskyi2020adversarial}) is one example to solve this problem, however, it often suffers from inferior generation quality (compared with original StyleGANs) and inaccurate reconstruction. We hypothesize that this is mainly because ALAE is trained to reconstruct fake images rather than real ones. Inspired by the success of in-domain GAN inversion~\cite{zhu2020domain}, we propose a novel algorithm to train a style encoder jointly with StyleGAN generator to reconstruct real images during GAN training. We term our method AE-StyleGAN.
%to learn the mapping from image space to $\mathcal{W}$ or $\mathcal{W}^+$ space. 

Another important aspect of autoencoders is their ability to learn disentangled representations. As stated in StyleGAN2~\cite{karras2020analyzing}, a more disentangled generator is easier to invert. This motivates us to ask if its inverse proposition is also true. We thus interpret our AE-StyleGAN objective as a regularization and explore: whether enforcing the generator to reconstruct real data (easy to invert) leads to more disentangled latent space.

Finally, we list our contribution as follows: (1) We propose AE-StyleGAN as improved techniques for training style-based autoencoders; (2) We discovered that an easy-to-invert generator is also more disentangled; (3) AE-StyleGAN shows superior generation and reconstruction quality than baselines.
% \begin{itemize}
%     \item We propose AE-StyleGAN as improved techniques for training style-based autoencoders.
%     \item We discovered that an easy-to-invert generator is also more disentangled.
%     \item AE-StyleGAN shows superior generation and reconstruction quality than baselines.
% \end{itemize}

\section{Related Works}
\noindent \textbf{GAN inversion.} The AE-StyleGAN is closely related to GAN inversion methods. However, we aim to solve a different problem: GAN inversion~\cite{abdal2019image2stylegan,DBLP:journals/corr/abs-1911-11544,zhu2020domain,alaluf2021restyle} aims at finding the most accurate latent code for the input image to reconstruct the input image with a pretrained and fixed GAN generator. While AE-StyleGAN aims to train an autoencoder-like structure where the encoder and generator are optimized end-to-end.
There are two streams of GAN inversion approaches: (1) to optimize the latent codes directly~\cite{abdal2019image2stylegan,DBLP:journals/corr/abs-1911-11544}, or (2) to train an amortised encoder to predict the latent code given images~\cite{zhu2020domain,alaluf2021restyle,richardson2021encoding,tov2021designing}.
%A suitable GAN inversion method should reconstruct the target image at the pixel level and align the inverted latent code with the semantic information encoded in the latent space.
% These methods take a pretrained generator and train an encoder or optimize the latents directly.
% Notably, in-domain GAN inversion~\cite{zhu2020domain} differs from traditional GAN inversion methods ~\cite{DBLP:journals/corr/abs-1906-08090, DBLP:journals/corr/PerarnauWRA16} where we first synthesizes a collection of images with randomly sampled latent codes and then uses the images and codes as inputs and supervisions respectively to train a deterministic model which misses semantic knowledge of data being modelled where as in in-domain GAN inversion a well-trained domain-guided encoder is included as a regularizer to land the latent code in the semantic domain during the optimization process which includes semantic knowledge in the inversion process. 
Notably, in-domain GAN inversion~\cite{zhu2020domain} proposes domain-guided encoder training that differs from traditional GAN inversion methods~\cite{DBLP:journals/corr/abs-1906-08090,DBLP:journals/corr/PerarnauWRA16} where encoders are trained with sampled image-latent pairs.

\noindent \textbf{Learning a bidirectional GAN.} There are two main approaches for learning a bidirectional GAN: (1) adversarial feature learning (BiGAN)~\cite{donahue2016adversarial} or adversarially learned inference (ALI)~\cite{dumoulin2016adversarially}; (2) combining autoencoder training with GANs, \eg VAE/GAN~\cite{larsen2016autoencoding}, AEGAN~\cite{lazarou2020autoencoding}. We focus on the latter since it usually gives better reconstruction quality and its training is more stable~\cite{tian2018cr}. Our AE-StyleGAN is different from these methods because these bidirectional models encode images in $\mathcal{Z}$ space (usually Gaussian) and are not designed for style-based generator networks.

\noindent \textbf{Adversarial Latent Autoencoders.} ALAE~\cite{pidhorskyi2020adversarial} is the most relevant work to ours. The proposed AE-StyleGAN is different from ALAE in that: (1) an ALAE discriminator is defined in latent space ($\mathcal{W}$) while ours is in image space; (2) ALAE reconstructs fake images by minimizing L2 between sampled $w$ and encoded fake image, while ours reconstruct real images.

\section{Method}
\subsection{Background}
\noindent \textbf{GAN and StyleGAN.} We denote image data as $\{x_i\}^n_{i=1} \subseteq \mathcal{X}$ drawn from the data distribution $P_{X}$. A generator is trained to transform samples $z \sim P_Z$ from a canonical distribution conditioned on labels to match the real data distributions. {\em Real} distributions are denoted as $P$ and {\em generated} distributions are denoted as $Q$. In this paper, we focus on StyleGANs and denote mapping network as $F: \mathcal{Z} \mapsto \mathcal{W}$, generator as $G: \mathcal{W} \mapsto \mathcal{X}$, and encoder as $E: \mathcal{X} \mapsto \mathcal{W}$ (or $\mathcal{W}^+$). We also denote fake and reconstructed images as $\tilde{x}=G(F(z))$ and $\hat{x}=G(E(x))$. The value function of GAN can be written as:
\begin{align}
    V_\text{GAN}(G \circ F,D) = &-\mathbb{E}_{x \sim P_{X}}{\mathcal{A}(-\tilde{D}(x))} \nonumber \\
    &-\mathbb{E}_{z \sim P_Z}{\mathcal{A}(\tilde{D}(G(F(z))))} \label{eq:gan}
\end{align}
\noindent Here $\mathcal{A}$ is the {\em activation function} and $\tilde{D}$ is the {\em logit} or discriminator's output before activation. Note that choosing $\mathcal{A}(t)=\text{softplus}(t)=\log{(1+\exp{(t)})}$ recovers the original GAN formulation~\cite{goodfellow2014generative,karras2019style}, and the resulting objective minimizes the Jensen-Shannon (JS) divergence between real and generated data distributions.

\noindent \textbf{In-domain GAN inversion.} In-domain GAN inversion~\cite{zhu2020domain} aims to learn a mapping from images to latent space. The encoder is trained to reconstruct real images (thus are ``in-domain'') and guided by image-level loss terms, \ie pixel MSE, VGG perceptual loss, and discriminator loss:
% \begin{align}
%     L_\text{idinv} =& \mathbb{E}_{x \sim P_{X}} [ \Vert x-G(E(x)) \Vert_2 + \nonumber \\
%     & \lambda_\text{vgg} \Vert h(x)-h(G(E(x))) \Vert_2 - \nonumber \\
%     & \lambda_\text{adv} \log{D(G(E(x)))} ], \label{eq:idinv}
% \end{align}
\begin{align}
    L_\text{idinv}(E,D,G) =& \mathbb{E}_{x \sim P_{X}} [ \Vert x-G(E(x)) \Vert_2 \nonumber \\
    &+\lambda_\text{vgg} \Vert h(x)-h(G(E(x))) \Vert_2 \nonumber \\
    &-\lambda_\text{adv} \mathcal{A}({-\tilde{D}(G(E(x)))}) ], \label{eq:idinv}
\end{align}
\noindent where $h$ is perception network and here we keep the same as in-domain inversion as VGG network.
% \ligong{fill things here}

\noindent \textbf{Negative data augmentation.} NDA~\cite{sinha2021negative} produces out-of-distribution samples lacking the typical structure of natural images. NDA-GAN directly specifies what the generator should not generate through NDA distribution $\Bar{P}$ and the resulting adversarial game is:
\begin{align}
    V_\text{NDA}(\lambda G \circ F + (1-\lambda) \Bar{P},D) \label{eq:nda}
\end{align}
\noindent where hyperparameter $\lambda$ is the mixture weight.

\subsection{Auto-Encoding StyleGANs}
\begin{algorithm}[h]
\caption{Decoupled AE-StyleGAN Training}
\label{alg:aegan_decouple}
\begin{algorithmic}[1]
    \State $\theta_E, \theta_{D_1}, \theta_{D_2}, \theta_F, \theta_G \leftarrow \text{Initialize network parameters}$
    \While{not converged}
        \State $x \leftarrow \text{Random mini-batch from dataset}$
        \State $z \leftarrow \text{Samples from } \mathcal{N}(0,I)$
        \State Step I. Update $D_1$, $D_2$
        \State $L_{D} \leftarrow -V_\text{GAN}(G \circ F,D_1)-V_\text{GAN}(G \circ E,D_2)$ in \ref{eq:gan}
        % \State $L_{D_2} \leftarrow -V(G \circ E,D_2)$ in \ref{eq:aegan}
        \State $\theta_{D_1}, \theta_{D_2} \leftarrow \text{ADAM}(\nabla_{\theta_{D_1},\theta_{D_2}} L_D, \theta_{D_1},\theta_{D_2})$
        \State Step II. Update $E$
        \State $L_E \leftarrow L_\text{idinv}(E,{D_2},G)$ in \ref{eq:idinv} or \ref{eq:idinv_ada}
        \State $\theta_E \leftarrow \text{ADAM}(\nabla_{\theta_E} L_E, \theta_E)$
        \State Step III. Update $F$, $G$
        \State $L_G \leftarrow V_\text{GAN}(G \circ F,{D_1})$ in \ref{eq:gan}
        \State $\theta_F, \theta_G \leftarrow \text{ADAM}(\nabla_{\theta_F,\theta_G} L_G, \theta_F,\theta_G)$
    \EndWhile
\end{algorithmic}
\end{algorithm}
\begin{algorithm}[h]
\caption{Joint AE-StyleGAN Training}
\label{alg:aegan_joint}
\begin{algorithmic}[1]
    \State $\theta_E, \theta_D, \theta_F, \theta_G \leftarrow \text{Initialize network parameters}$
    \While{not converged}
        \State $x \leftarrow \text{Random mini-batch from dataset}$
        \State $z \leftarrow \text{Samples from } \mathcal{N}(0,I)$
        \State Step I. Update $D$
        \State $L_D \leftarrow -V_\text{AEGAN}(G \circ F,D)$ in \ref{eq:aegan}
        \State $\theta_D \leftarrow \text{ADAM}(\nabla_{\theta_D} L_D, \theta_D)$
        \State Step II. Update $E$, $G$
        \State $L_E \leftarrow L_\text{idinv}(E, D, G)$ in \ref{eq:idinv} or \ref{eq:idinv_ada}
        \State $\theta_E, \theta_G \leftarrow \text{ADAM}(\nabla_{\theta_E,\theta_G} L_E, \theta_E,\theta_G)$
        \State Step III. Update $F$, $G$
        \State $L_G \leftarrow V_\text{AEGAN}(G \circ F,D)$ in \ref{eq:aegan}
        \State $\theta_F, \theta_G \leftarrow \text{ADAM}(\nabla_{\theta_F,\theta_G} L_G, \theta_F,\theta_G)$
    \EndWhile
\end{algorithmic}
\end{algorithm}

\noindent \textbf{Decoupled AE-StyleGAN.} One straightforward way to train encoder and generator end-to-end is to simultaneously train an encoder with GAN inversion algorithms along with the generator. Here we choose in-domain inversion. To keep the generator's generating ability intact, one can decouple GAN training and GAN inversion training by introducing separate discriminator models $D_1$ and $D_2$, and freezing $G$ in inversion step. Specifically, $D_1$ is involved in $V_\text{GAN}(G \circ F, D_1)$ in Equation~\ref{eq:gan} for GAN training steps, and $D_2$ is involved in $L_\text{idinv}(E,D_2,G)$ in Equation~\ref{eq:idinv}. Training of $D_2$ follows:
\begin{align}
    V_\text{GAN}(G \circ F,D_2) = &-\mathbb{E}_{x \sim P_{X}}{\mathcal{A}(-\tilde{D}_2(x))} \nonumber \\
    &-\mathbb{E}_{x \sim P_X}{\mathcal{A}(\tilde{D}_2(G(E(x))))}. \label{eq:gan_2}
\end{align}
\noindent Please refer to Figure~\ref{fig:arch}-a and Algorithm~\ref{alg:aegan_decouple} for details.

It is worth mentioning that this decoupled algorithm is similar to CR-GAN~\cite{tian2018cr} except that we use decoupled discriminators.
\begin{figure*}[t]
\begin{center}
%   \includegraphics[width=0.8\linewidth]{wacv-records/Recon-4x4-Grids/Wplus-sample-grid-4-datasets.pdf}
    \includegraphics[width=0.85\linewidth]{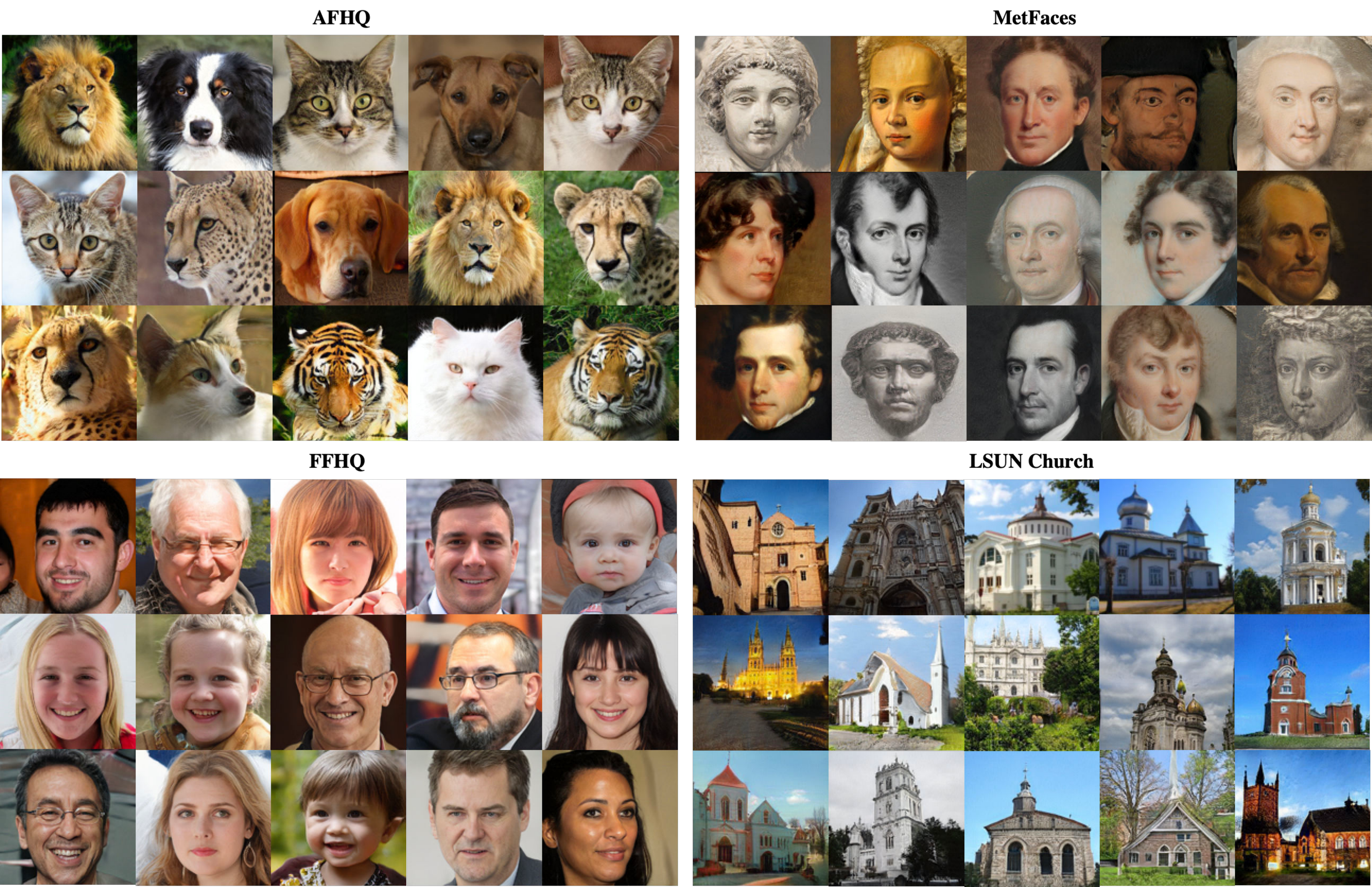}
\end{center}
   \caption{Generations with AE-StyleGAN ($\mathcal{W}^+$). The model generates lot of diverse images which is justified by its LPIPS score tabulated in Table~\ref{tab:fid_lpips}.}
% \label{fig:long}
% \label{fig:onecol}
% \label{fig:wplusGen}
\label{fig:sample_aegan}
\end{figure*}
\begin{figure*}[t]
\begin{center}
  \includegraphics[width=0.85\linewidth]{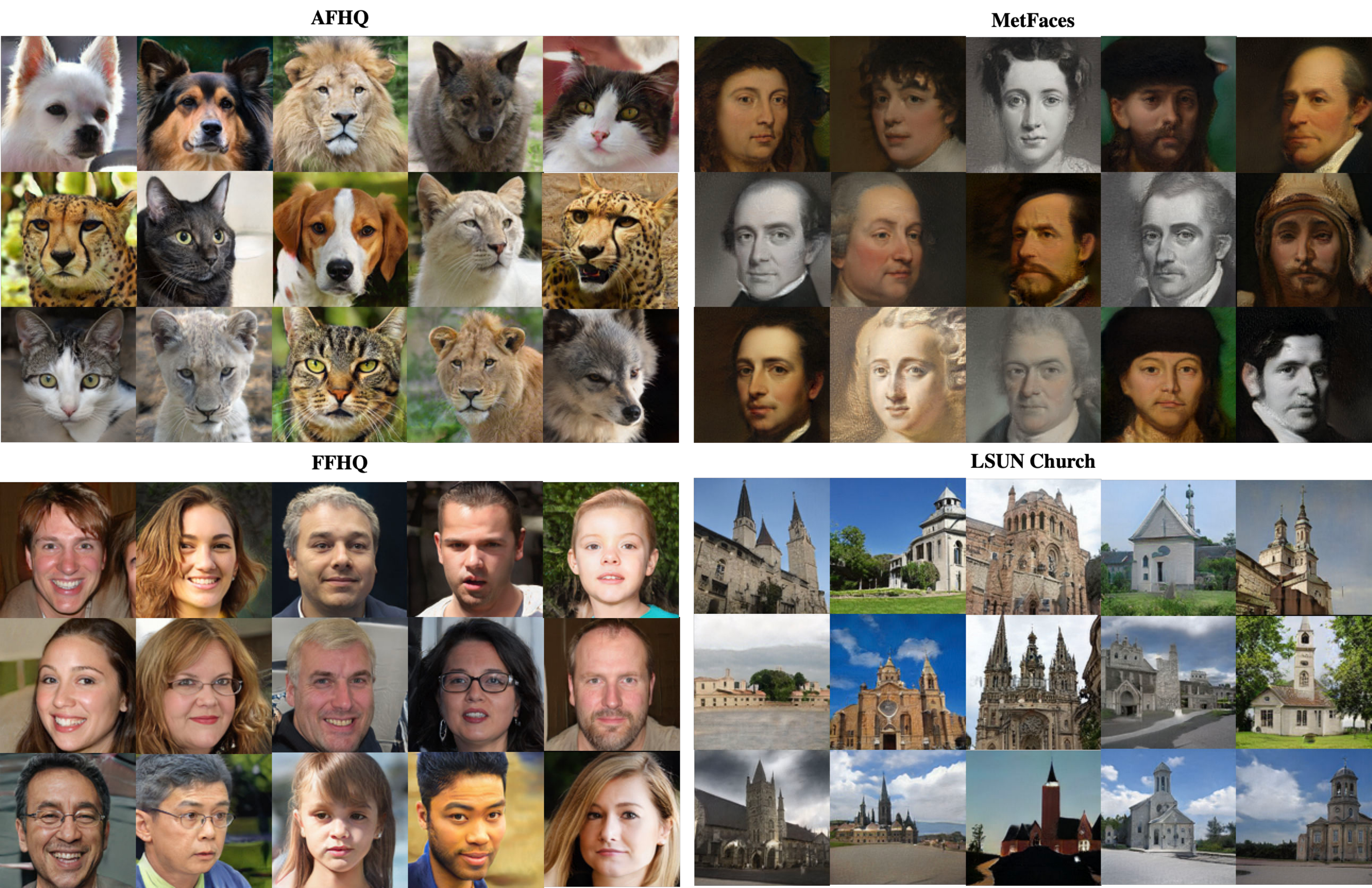}
\end{center}
   \caption{Generations with ALAE. Although, these are comparable in quality, AE-StyleGAN ($\mathcal{W}^+$) still outperforms ALAE by huge margin Table~\ref{tab:fid_lpips}}
% \label{fig:long}
% \label{fig:onecol}
% \label{fig:ALAEGen}
\label{fig:sample_alae}
\end{figure*}
% \ligong{please make above Figure 3 (fig:sample\_alae) same as Figure 2, i.e. make them have same margin and white spaces}

\begin{figure*}[t]
\begin{center}
    %  \includegraphics[width=0.8\linewidth]{wacv-records/Recon-4x4-Grids/Wplus-recon-grid-4-datasets.pdf}
    \includegraphics[width=0.95\linewidth]{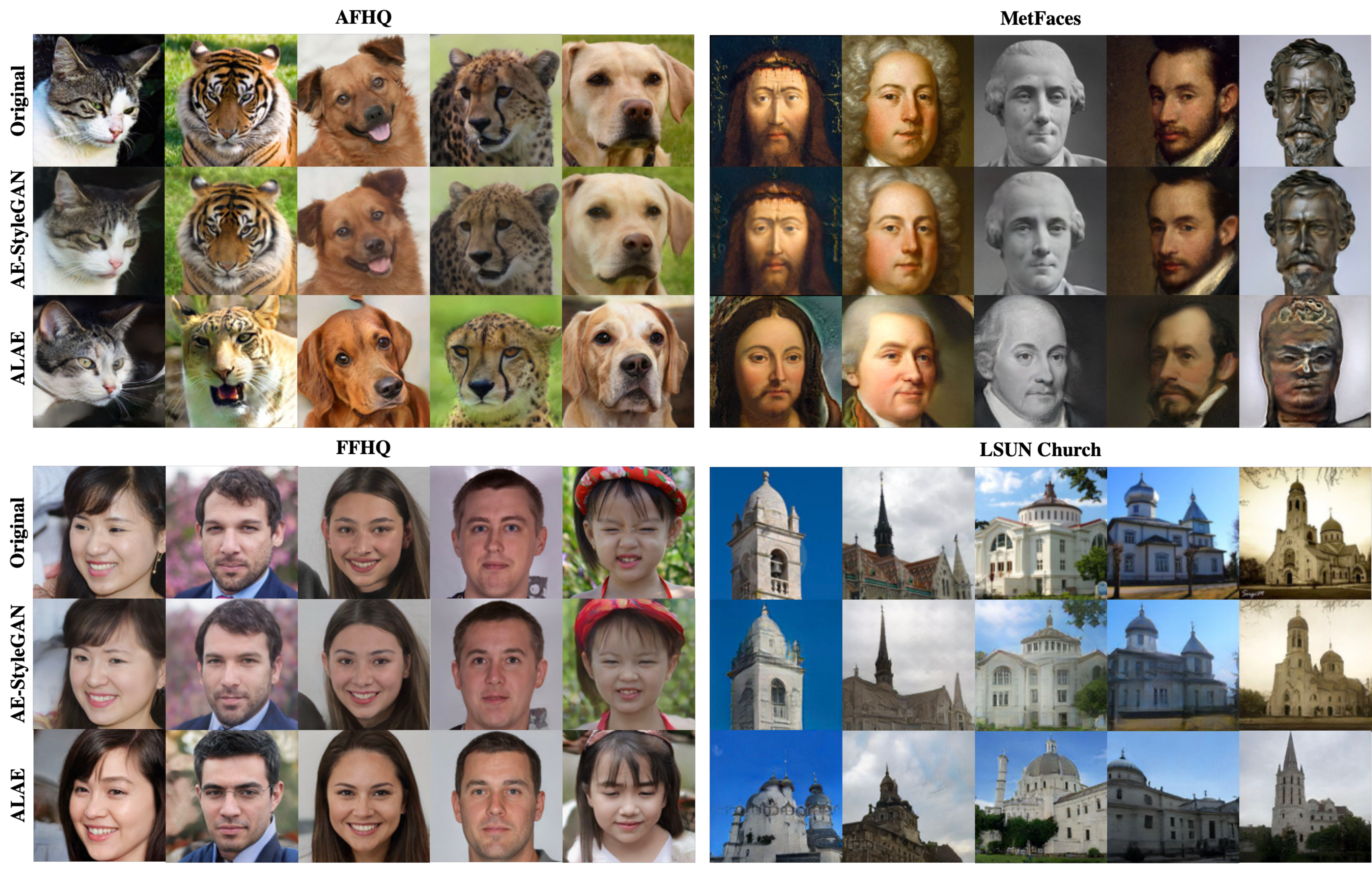}
\end{center}
   \caption{Reconstructions with AE-StyleGAN ($\mathcal{W}^+$) and ALAE. AE-StyleGAN not only reconstructs the main objects but also preserves details like background, color, expression, \etc. While reconstructions with ALAE suffer with identity loss. For example, the Dog and Tiger images from AFHQ in this grid are reconstructed with modified expression, face shape, color.}
% \label{fig:long}
% \label{fig:onecol}
% \label{fig:wplusRecon}
\label{fig:recon_aegan_alae}
\end{figure*}
% \ligong{please merge fig:recon\_alae into above Figure 4 fig:recon\_aegan\_alae}
% \begin{figure*}[t]
% \begin{center}
% % \fbox{\rule{0pt}{2in} \rule{0.9\linewidth}{0pt}}
% %   \includegraphics[width=0.8\linewidth]{wacv-records/Recon-4x4-Grids/ALAE-recon-grid-4-datasets.pdf}
%     \includegraphics[width=0.95\linewidth]{wacv-records/Recon-4x4-Grids/ALAE-recon-grid-4-datasets-2x5.pdf}
% \end{center}
%   \caption{Reconstructions with ALAE suffer with identity loss. For example, the Dog and Tiger images from AFHQ in this grid are reconstructed with modified expression, face shape, color, etc. and are far from the originals.}
% % \label{fig:long}
% % \label{fig:onecol}
% % \label{fig:ALAERecon}
% \label{fig:recon_alae}
% \end{figure*}

\noindent \textbf{Joint AE-StyleGAN.} The generator of a decoupled AE-StyleGAN would be exactly equivalent to a standard StyleGAN generator, however, we often find the encoder not capable of faithfully reconstruct real images. This phenomenon is illustrated in Figure~\ref{fig:decouple_vs_joint}. We hypothesize that with $G$ frozen at inversion step, $E$ cannot catch up with $G$'s update, thus lags behind $G$. To cope with this issue, we propose to train $G$ jointly with $E$ in the inversion step. We also use a single discriminator for both pathways. For the GAN pathway, the value function is written as:
\begin{align}
    V_\text{AEGAN}(G \circ F,D) = &-\mathbb{E}_{x \sim P_{X}}{\mathcal{A}(-\tilde{D}(x))} \label{eq:aegan} \\
    &-\lambda_\text{adv} \mathbb{E}_{x \sim P_{X}}{\mathcal{A}(\tilde{D}(G(E(x))))} \nonumber \\
    &-(1-\lambda_\text{adv}) \mathbb{E}_{z \sim P_Z}{\mathcal{A}(\tilde{D}(G(F(z))))}. \nonumber
\end{align}
This can be viewed as a differentiable NDA, with $\Bar{P}$ set to be the distribution of reconstructed real images, and $\lambda=1-\lambda_\text{adv}$. Please refer to Figure~\ref{fig:arch}-b and Algorithm~\ref{alg:aegan_joint} for details. In the following text, we use AE-StyleGAN to refer Joint AE-StyleGAN, if not explicitly specified.

\noindent \textbf{Adaptive discriminator weight.} Inspired by VQGAN~\cite{esser2021taming}, we also experimented with adding an adaptive weight to automatically balance reconstruction loss and adversarial loss:
\begin{align}
    L^\text{ada}_\text{idinv} =& L_\text{rec} + \beta L_\text{adv}, \nonumber \\
    \text{where}~~L_\text{rec} =& \mathbb{E}_{x \sim P_{X}} [ \Vert x-G(E(x)) \Vert_2 \nonumber \\
    &+\lambda_\text{vgg} \Vert h(x)-h(G(E(x))) \Vert_2 ], \nonumber \\
    \text{and}~~ L_\text{adv} =& -\lambda_\text{adv} \mathbb{E}_{x \sim P_{X}} [ \mathcal{A}({-\tilde{D}(G(E(x)))}) ], \nonumber \\
    \text{with}\quad~~ \beta =& \frac{\nabla_{\theta^l_{G}}[L_\text{rec}]}{\nabla_{\theta^l_{G}}[L_\text{adv}]+\epsilon},\label{eq:idinv_ada}
\end{align}
\noindent where $L_\text{rec}$ is the weighted sum of pixel reconstruction loss and VGG perceptual loss, $\nabla_{\theta^l_{G}}[\cdot]$ is the gradient of its input w.r.t. the last layer $l$ of generator, and $\epsilon=10^{-6}$ added for numerical stability. In the following text, we use AE-StyleGAN with adaptive $\beta$ for all experiments if not specified.

%------------------------------------------------------------------------
\begin{table*}[h!]
    %\vspace{-0.05in}
    \caption{\textbf{Fr\'{e}chet Inception Distances (FID)}, and the \textbf{Learned Perceptual Image  Patch  Similarity (LPIPS)} are computed for all the models over all datasets. Our proposed Joint AE-StyleGAN achieves the lowest FID and highest LPIPS in most cases. The top-two best performing methods are marked in boldface.}
    % \vspace{0.1in}
    \label{tab:compare}
    \centering
    \scalebox{1}{
    % \resizebox{1\textwidth}{!}{
    \begin{tabular}{lccccccccc}
    %\hline
    \toprule
    % \multicolumn{1}{l}{} & \multicolumn{8}{c}{{PPL}} \\
    % \cmidrule(lr){2-9}
    \multicolumn{1}{l}{} & \multicolumn{2}{c}{{StyleGAN}} & \multicolumn{2}{c}{{ALAE}} & \multicolumn{2}{c}{{AE-StyleGAN} ($\mathcal{W}$)} & \multicolumn{2}{c}{{AE-StyleGAN} ($\mathcal{W}^+$)}\\
    \cmidrule(lr){2-3} \cmidrule(lr){4-5} \cmidrule(lr){6-7} \cmidrule(lr){8-9} %\cmidrule(lr){14-16}
    & FID $\downarrow$ & LPIPS $\uparrow$ & FID $\downarrow$ & LPIPS $\uparrow$ & FID $\downarrow$ & LPIPS $\uparrow$ & FID $\downarrow$ & LPIPS $\uparrow$\\
    \hline
    FFHQ &  {\bf 7.359} & 0.432   &  12.574 & 0.438 & 8.176 & {\bf 0.448}  & {\bf 7.941}	& {\bf 0.451}\\
    AFHQ &  {\bf 7.992} & 0.496  & 21.557 & 0.508 & 15.655 & {\bf 0.522}  &  {\bf 10.282} & {\bf 0.518}\\
    MetFaces &  {\bf 29.318} & 0.465  & 41.693 & 0.462 & 29.710 & {\bf 0.469} &  {\bf 29.041} & {\bf 0.471}\\
    LSUN Church &  {\bf 27.780} & 0.520  & 29.999 & 0.552 & 29.387 & {\bf 0.603}  &  {\bf 29.358} & {\bf 0.592}\\
    % \hline
    % Average Rank & 3.5 & 4 & 3 & 4 & 3.75 & 5 & 3.5 & 3.25 & 2.75 & {\bf 1.75} &  2.5 & 2.5\\ %& 2.25 & {\bf 1.5} & {\bf 1.75} \\
    \bottomrule
    \end{tabular}}
    \label{tab:fid_lpips}
\end{table*}
\begin{table*}[h]
    %\vspace{-0.05in}
    \caption{\textbf{PPL}. Perceptual path lengths on FFHQ, AFHQ, MetFaces and Church dataset are computed on all models over all datasets (lower is better). Our model is performing 13\% to 20\% better than ALAE on complex datasets like LSUN church and FFHQ. The top-two best performing methods are marked in boldface. }
    % \vspace{0.1in}
    \label{tab:compare}
    \centering
    \scalebox{1}{
    % \resizebox{1\linewidth}{!}{
    \begin{tabular}{lccccccccc}
    %\hline
    \toprule
    \multicolumn{1}{l}{} & \multicolumn{2}{c}{{StyleGAN}} & \multicolumn{2}{c}{{ALAE}} & \multicolumn{2}{c}{{AE-StyleGAN} ($\mathcal{W}$)} & \multicolumn{2}{c}{{AE-StyleGAN} ($\mathcal{W}^+$)}\\
    \cmidrule(lr){2-3} \cmidrule(lr){4-5} \cmidrule(lr){6-7} \cmidrule(lr){8-9} %\cmidrule(lr){14-16}
    & Full & End & Full & End & Full & End & Full & End\\
    \hline
    FFHQ &  {\bf 173.09} & {\bf 173.68}   &  192.60 & 193.94 & 181.03  &  180.23  & {\bf 166.70} & {\bf 165.85} \\
    AFHQ &  244.83 & 240.68  & {\bf 229.54} & {\bf 232.53} & 247.75	& 248.75  &  {\bf 233.86} & {\bf 231.91} \\
    MetFaces &  {\bf 231.40} & {\bf 232.77} & {\bf 235.39} & 237.63  &  238.84 & 238.60 & 240.01	& {\bf 235.41} \\
    LSUN Church & 245.22 & {\bf 239.62}  & 298.06 & 295.01 & {\bf 241.46} & 231.73 & {\bf 240.01}	& {\bf 231.49} \\
    % \hline
    % Average Rank & 3.5 & 4 & 3 & 4 & 3.75 & 5 & 3.5 & 3.25 & 2.75 & {\bf 1.75} &  2.5 & 2.5\\ %& 2.25 & {\bf 1.5} & {\bf 1.75} \\
    \bottomrule
    \end{tabular}}
    \label{tab:ppl}
\end{table*}

\begin{table*}[h]
    %\vspace{-0.05in}
    \caption{\textbf{Reconstruction VGG, Per Pixel MSE and FID comparison.} VGG perceptual loss, Pixel MSE and FID values are computed between real images and their reconstructions for ALAE, Joint AE-StyleGAN ($\mathcal{W}$) \& Joint AE-StyleGAN ($\mathcal{W}^+$). The best performing methods are marked in boldface.}
    \label{tab:compare}
    \centering
    \scalebox{1}{
    % \resizebox{1\linewidth}{!}{
    \begin{tabular}{lcccccccccc}
    %\hline
    \toprule
    \multicolumn{1}{l}{}  & \multicolumn{3}{c}{{ALAE}} & \multicolumn{3}{c}{{AE-StyleGAN} ($\mathcal{W}$)} & \multicolumn{3}{c}{{AE-StyleGAN} ($\mathcal{W}^+$)}\\
    \cmidrule(lr){2-4} \cmidrule(lr){5-7} \cmidrule(lr){8-10}
    & VGG $\downarrow$ & MSE $\downarrow$ & FID $\downarrow$ & VGG$\downarrow$ & MSE $\downarrow$ & FID $\downarrow$ & VGG $\downarrow$ & MSE $\downarrow$ & FID $\downarrow$ \\
    \hline
    FFHQ & 0.81 & 64.82 & 21.68 & 0.28 & 26.10 & 16.76 & {\bf 0.26 }& {\bf 25.34} & {\bf 14.67} \\
    AFHQ & 0.99 & 73.77 & 25.30 & 0.29 & 29.21	& 5.37  & {\bf 0.27 }& {\bf 28.75} & {\bf 4.90} \\
    MetFaces & 0.54 & 49.72 & 52.55 & 0.05 & 13.48 & 27.38 & {\bf 0.05 }& {\bf 13.34} & {\bf 25.34} \\
    LSUN Church & 0.86 & 76.97 & 43.52 & 0.30 & 32.29 & {\bf 26.45} & {\bf 0.29 }& {\bf 31.75} & 32.67 \\
    \bottomrule
    \end{tabular}}
    \label{tab:vgg_mse_fid}
\end{table*}

\section{Experiments}
%-------------------------------------------------------------------------
\noindent \textbf{Datasets.}
We evaluate AE-StyleGAN with four datasets: (1) FFHQ dataset which consists of 70,000 images of people faces aligned and cropped at resolution of $1024 \times 1024$; (2) AFHQ dataset consisting of 15,000 images from domains of cat, dog, and wildlife; (3) MetFaces dataset consists of 1,336 high-quality human face images at $1024 \times 1024$ resolution collected from art works in the Metropolitan Museum; (4) The LSUN Churches contains 126,000 outdoor photographs of churches of diverse architectural styles. For a fair comparison, we resized all images to $128 \times 128$ resolution for training.

\noindent \textbf{Baselines.} We compare our proposed models with ALAE as a competitive baseline. For a fair comparison, we implemented ALAE following the paper~\cite{DBLP:journals/corr/abs-2004-04467}, Joint AE-StyleGAN, Decoupled AE-StyleGAN on the same backbone StyleGAN2. %The code is written in PyTorch~\cite{paszke2019pytorch}.

\noindent \textbf{Evaluation metrics.} Fr\'{e}chet Inception Distances (FID)~\cite{heusel2017gans}, learned perceptual image patch similarity (LPIPS), Perceptual Path Length (PPL) and are reported for quantitative evaluation in Table~\ref{tab:fid_lpips} and Table~\ref{tab:ppl}. Experimental setup and additional results are detailed in Appendix.
% We evaluate AE-StyleGAN on FFHQ [5], AFHQ[6], MetFaces[7], Church[8]. All images are resized to
% 128 × 128 resolution for training.

\subsection{Implementation}
% \noindent \textbf{Encoder architecture.} 
The code is written in PyTorch~\cite{paszke2019pytorch} and heavily based on Rosinality's implementation\footnote{\url{https://github.com/rosinality/stylegan2-pytorch}} of StyleGAN2~\cite{karras2020analyzing}.
For image encoders, we modify the last linear layer of the image discriminator to desired dimensions, and remove its mini-batch standard deviation layer. We keep the default values for hyperparameters of in-domain inversion and fix $\lambda_\text{vgg}=5\times 10^{-5}$ and $\lambda_\text{adv}=0.1$ for all experiments. For a fair comparison, we reimplemented ALAE under the same codebase.

%-------------------------------------------------------------------------
\subsection{Qualitative Analysis}
\begin{figure}[t]
\begin{center}
% \fbox{\rule{0pt}{2in} \rule{.9\linewidth}{0pt}}
\includegraphics[width=1\linewidth]{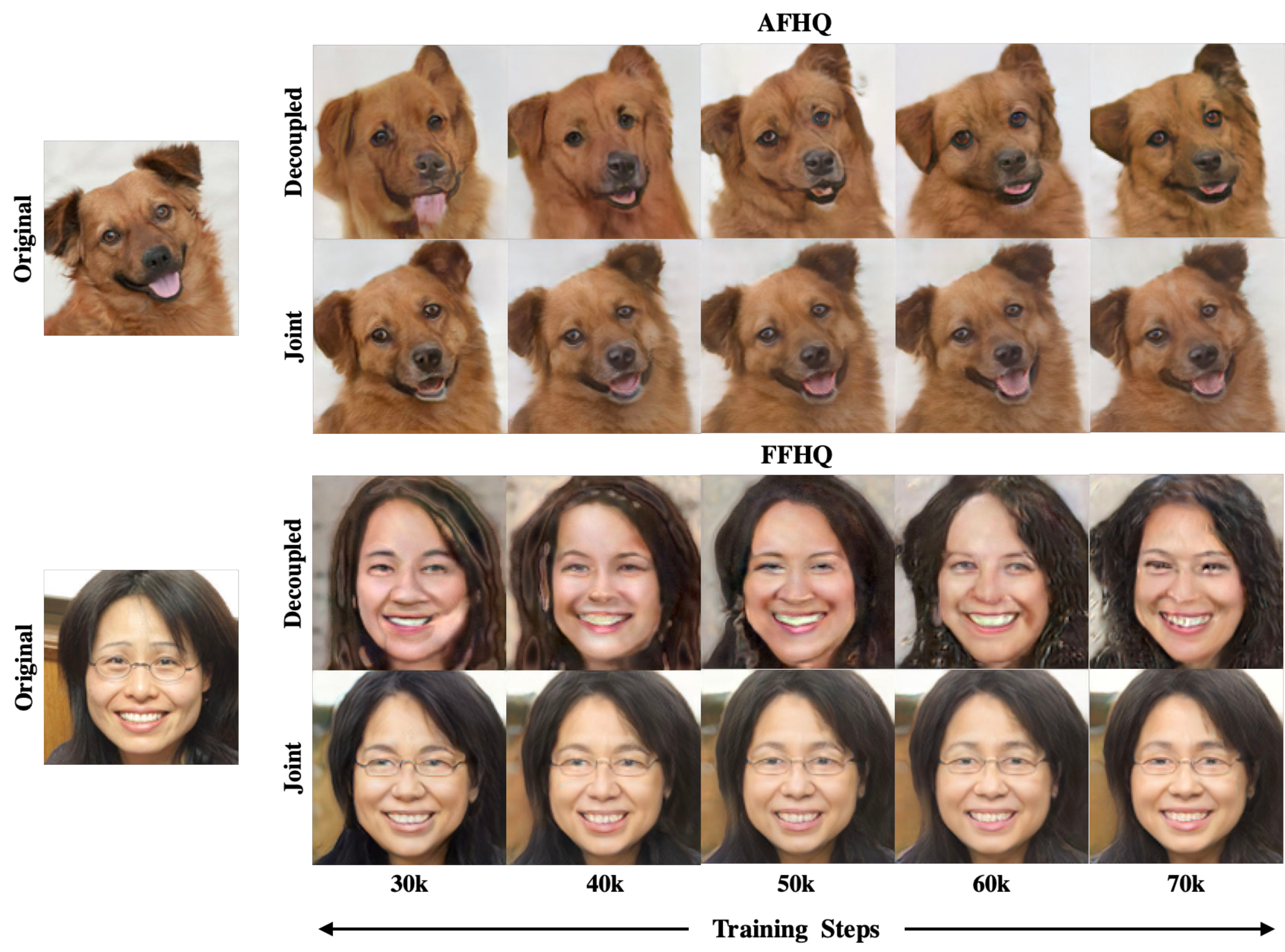}
\end{center}
   \caption{Reconstruction progress for AFHQ and FFHQ for Decoupled and Joint AE-StyleGAN. Decoupled model's reconstructions are with noise, artifacts, identity loss. Joint AE-StyleGAN faithfully reconstructs the image as the training progresses.}
\label{fig:decouple_vs_joint}
\end{figure}

\noindent \textbf{Decoupled \vs Joint AE-StyleGAN.}
We also compare Decoupled AE-StyleGAN with Joint AE-StyleGAN with a visualization of real image reconstruction as the training progresses.  From Figure~\ref{fig:decouple_vs_joint}, especially for a complex dataset like FFHQ, we see a lot of noise, artifacts in the reconstructions of Decoupled AE-StyleGAN which can be explained with the fact that the encoder is not being trained jointly with generator, hence it is failing to cope up with generator's learning curve thus compromising on the reconstruction quality. However, as the encoder is being trained jointly with generator in Joint AE-StyleGAN, it does not lag behind generator which helps it to faithfully reconstruct the real image. Another interesting observation to support this argument is that the Decoupled AE-StyleGAN's encoder is performing much better on AFHQ data which is a simple dataset when compared with FFHQ. However, we can observe strong facial deformation and Jointly trained encoder in Joint AE-StyleGAN is able to reconstruct it well unanimously. This analysis also adds strong weight on our claim that training an encoder jointly with generator helps in finding a more disentangled latent space thus, reinforcing a better real image reconstruction capability.

\noindent \textbf{ALAE \vs Joint AE-StyleGAN.}
We compare Joint AE-StyleGAN with ALAE through sampled images and real image reconstructions. From Figure~\ref{fig:sample_aegan} and Figure~\ref{fig:sample_alae} we can observe that the sample image generation quality of ALAE is comparable to Joint AE-StyleGAN qualitatively (Joint AE-StyleGAN still outperforms ALAE in terms of FID and LPIPS by a large margin, see Table~\ref{tab:fid_lpips}). However, ALAE fails to reconstruct a real image accurately, where the reconstruction looses its identity from the original image which is very evident from Figure~\ref{fig:recon_aegan_alae}. On the other hand, Joint AE-StyleGAN does not only reconstruct a real image faithfully but also preserves details like background, color, expression, \etc. outperforming ALAE as shown in Figure~\ref{fig:recon_aegan_alae}.

\begin{figure*}
\begin{center}
% \fbox{\rule{0pt}{2in} \rule{.9\linewidth}{0pt}}
\includegraphics[width=1\linewidth]{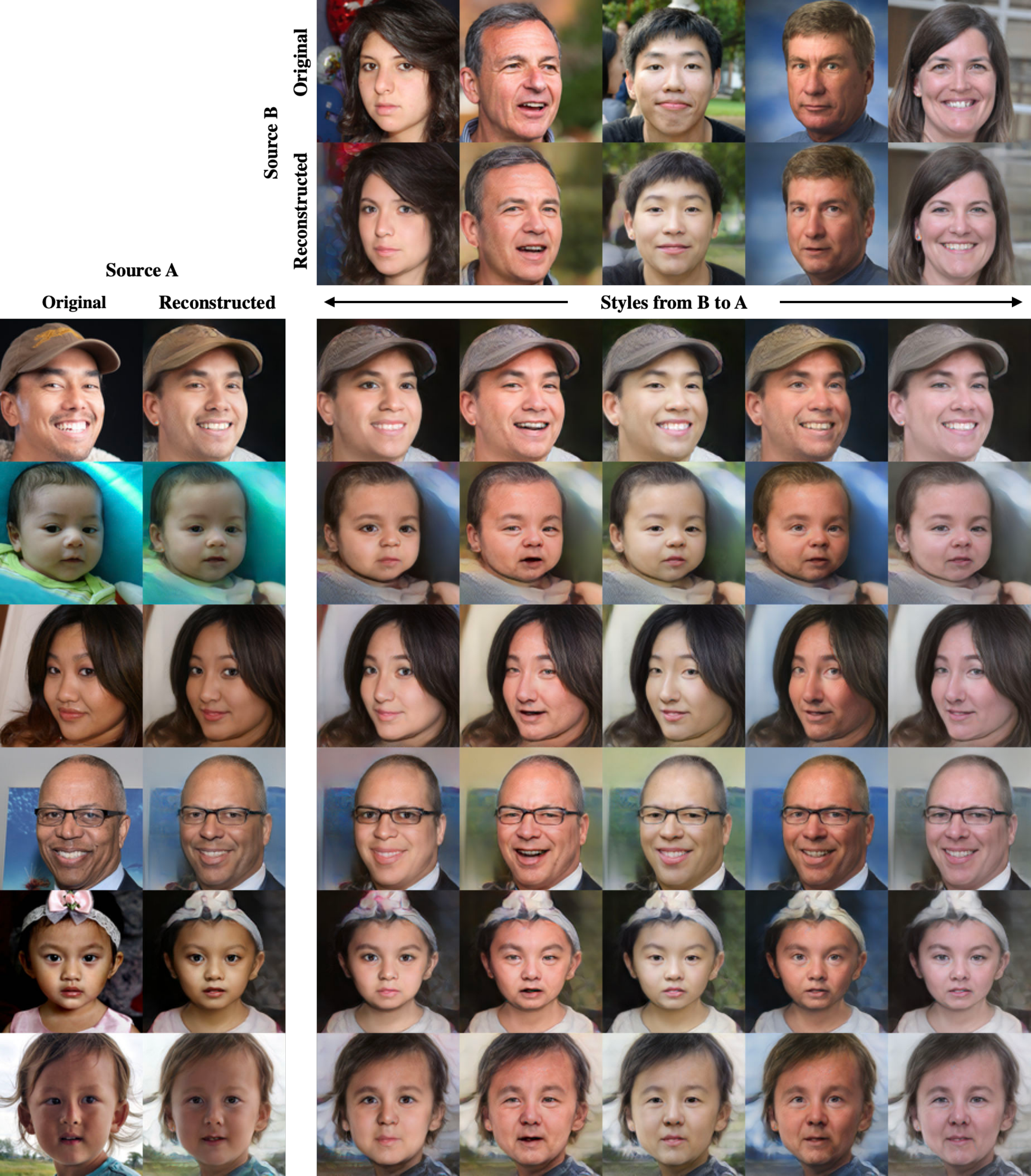}
\end{center}
   \caption{\textbf{Style transfer within FFHQ.} Style transfer experiment between two sets of real images from FFHQ dataset. We reconstruct real images via Joint AE-StyleGAN $\mathcal{W}^+$ and use $\mathcal{W}^+$ latent codes of all the images and copy specified subset of styles from source B to source A, thus creating an image grid as above that show original images and their reconstructions as well as the style mixed images from source B to source A}
\label{fig:stylemix}
\end{figure*}

\noindent \textbf{Style transfer.}
We also experimented with style transfer as shown in Figure~\ref{fig:stylemix} between real images from FFHQ to support our claim that Joint AE-StyleGAN's training algorithm in fact creates a more disentangled $\mathcal{W}$ (or $\mathcal{W}^+$) space. We use an encoder trained via Joint AE-StyleGAN ($\mathcal{W}^+$) method and pass 11 real images that are randomly chosen from FFHQ dataset to the encoder to obtain $\mathcal{W}^+$ latent codes. As we are working with $128 \times 128$ resolution images, our $\mathcal{W}^+$ latent code consists of 12 styles latent vectors, each of size 512. We experimented with various combinations of these latent vectors from source B to source A and found that mixing 7 to 11 style latents from source B's $\mathcal{W}^+$ latent code to source A's $\mathcal{W}^+$ latent code gave us meaningful style transfer results. Thus proving our argument that Joint AE-StyleGAN training methodology in fact helps in creating a more disentangled $\mathcal{W}$ (or $\mathcal{W}^+$) space and helps in real image editing.

% Table with ref_vgg, ref_pix, recon_fid comparison for ALAE and WPlus, WTied ---> make a table with the checkpoint of best FiD ---> for all datasets
% Table with all datasets and fid vs ppl vs lpips score comparision

% \begin{figure*}
% \begin{center}
% % \fbox{\rule{0pt}{2in} \rule{.9\linewidth}{0pt}}
% \includegraphics[width=0.8\linewidth]{wacv-records/stylemix/alae/ffhq-grid.png}
% \end{center}
%   \caption{stylemix-ffhq-alae.}
% \label{fig:short}
% \end{figure*}

% \begin{figure*}
% \begin{center}
% % \fbox{\rule{0pt}{2in} \rule{.9\linewidth}{0pt}}
% \includegraphics[width=0.8\linewidth]{wacv-records/stylemix/wplus/ffhq-samples-3style-grid.png}
% \end{center}
%   \caption{stylemix-ffhq-Wplus-Sample Images.}
% \label{fig:short}
% \end{figure*}

\subsection{Quantitative Analysis}
We did a quantitative comparison of ALAE, {AE-StyleGAN} ($\mathcal{W}$), {AE-StyleGAN} ($\mathcal{W}^+$) and StyleGAN2 by computing FID, LPIPS.
% to justify our claim that training an autoencoder jointly with the generator helps to map image space to a more disentangled latent space.
First, to be on the common ground, we use the same implementation of StyleGAN2 to serve as a backbone for ALAE and AE-StyleGAN. Upon optimising these models over four datasets FFHQ, AFHQ, MetFaces and LSUN Church we are reporting FID and LPIPS scores through Table~\ref{tab:fid_lpips}. Our model surpassed ALAE in terms of FID and LPIPS in all datasets. To strengthen the evidence, we also did PPL computations on every model for all the datasets. It is interesting to note from Table~\ref{tab:ppl} that our model beats ALAE in most cases. Especially, for complex datasets like FFHQ and LSUN Church, PPL scores of AE-StyleGAN is 13\% to 22\% better than those of ALAE. Although PPL scores of ALAE for simpler datasets like AFHQ ahd MetFaces is comparable, our model still outperforms ALAE.
At last, we compared VGG loss (or perceptual loss), Pixel loss, FID of real images and reconstructed images for all the best models and tabulated them at Table~\ref{tab:vgg_mse_fid}. We can also see that Joint AE-StyleGAN outperforms ALAE in terms of inverting the generator with an observed VGG, MSE losses more than 50\% less than ALAE.
% With vivid analysis both qualitatively and quantitatively, we can conclude that Joint AE-StyleGAN is better than ALAE.

% \ligong{please fix Figure 6}

\subsection{Ablation Study}
% \noindent \textbf{.} 
In our early experiments, we ablate hyperparameters including whether to use decoupled discriminators, whether to jointly update $G$, and the number of $E$ step per $G$ step on VoxCeleb2 dataset~\cite{chung2018voxceleb2} at resolution $64 \times 64$. Sample FID and per pixel reconstruction MSE are reported in Table~\ref{tab:ablation}. We can observe that if $G$ is not jointly updated with $E$, the reconstruction MSE only decreases slightly even with 4 $E$-steps. While jointly updating $G$ improves MSE substantially.
\begin{table}[h]
    \caption{Ablation studies. Decouple $D$: whether to use decoupled discriminators; Joint $G$: whether to jointly update $G$; $\#$ $E$-step: number of $E$ step per $G$ step. All metrics are evaluated at 20000 iterations.}
    \label{tab:ablation}
    \centering
    % \resizebox{1\linewidth}{!}{
    \scalebox{1}{
    \begin{tabular}{lllcc}
    \toprule
    {Decouple $D$}  & {Joint $G$}   & {$\#$ $E$-step}   & FID $\downarrow$ & MSE $\downarrow$ \\ \hline
    \cmark&\xmark&1 & 36.101 & 13.716 \\
    \cmark&\xmark&4 & 34.632 & 12.202 \\
    \cmark&\cmark&1 & 35.384 &  8.701 \\
    \cmark&\cmark&4 & 35.409 &  5.872 \\
    \xmark&\xmark&1 & 39.611 & 13.229 \\
    \xmark&\xmark&4 & 37.559 & 12.822 \\
    \xmark&\cmark&1 & 35.048 &  8.454 \\
    \xmark&\cmark&4 & 36.602 &  5.924 \\
    \bottomrule
    \end{tabular}}
\end{table}

\begin{table}[h!]
    \caption{\textbf{Adaptive discriminator weight.} Comparing sample FID and reconstruction MSE evaluated on FFHQ dataset. With adaptive weight $\beta$, generator achieves better FID and MSE.}
    \label{tab:compare}
    \centering
    \scalebox{1}{
    % \resizebox{1\linewidth}{!}{
    \begin{tabular}{lcccc}
    %\hline
    \toprule
    \multicolumn{1}{l}{} & \multicolumn{2}{c}{{AE-StyleGAN} ($\mathcal{W}$)} & \multicolumn{2}{c}{{AE-StyleGAN} ($\mathcal{W}^+$)}\\
    \cmidrule(lr){2-3} \cmidrule(lr){4-5}
     & w/o $\beta$ & w/ $\beta$ & w/o $\beta$ & w/ $\beta$\\
    \hline
    FID $\downarrow$ & 8.620 & 8.177 & 8.241 & {\bf 7.941} \\
    MSE $\downarrow$ & 30.683 & 26.107 & 29.726 & {\bf 25.341} \\
    \bottomrule
    \end{tabular}}
    \label{tab:ada}
\end{table}

% \noindent \textbf{Adaptive discriminator weight.} 
We also ablate adaptive discriminator weight defined in Equation~\ref{eq:idinv_ada} on FFHQ at $128 \times 128$ resolution. Sample FID and per pixel reconstruction MSE are reported in Table~\ref{tab:ada}. We can observe that using adaptive weight consistently improves both FID and MSE metrics.

% \section{Conclusion and Discussion}
\section{Conclusion}
%conclusion.
In this paper, we proposed AE-StyleGAN, a novel algorithm that jointly trains an encoder with a style-based generator. With empirical analysis, we confirmed that this methodology provides an easy-to-invert encoder for real image editing. Extensive results showed that our model has superior image generation and reconstruction capability than baselines.
We have explored the problem of training an end-to-end autoencoder. With improved generation fidelity and reconstruction quality, the proposed AE-StyleGAN model can serve as a building-block for further development and applications. For example, it could potentially improve CR-GAN~\cite{tian2018cr} where an encoder is involved in generator training. It also enables further improvement of disentanglement by borrowing techniques such as Factor-VAE~\cite{kim2018disentangling}. We leave these for future work.

% \newpage
{\small
\bibliographystyle{ieee_fullname}
\bibliography{refs}
}

\end{document}